\documentclass[10pt,twocolumn,letterpaper]{article}

\usepackage[pagenumbers]{cvpr} 

\usepackage{graphicx}
\usepackage{amsmath}
\usepackage{amssymb}
\usepackage{booktabs}

\usepackage{multirow}
\usepackage{multicol}
\usepackage{makecell}
\usepackage{amsmath}
\usepackage{amssymb}
\usepackage{color, xcolor}
\usepackage{colortbl}

\usepackage[pagebackref,breaklinks,colorlinks]{hyperref}

\usepackage[capitalize]{cleveref}
\crefname{section}{Sec.}{Secs.}
\Crefname{section}{Section}{Sections}
\Crefname{table}{Table}{Tables}
\crefname{table}{Tab.}{Tabs.}

\def\cvprPaperID{*****} 
\def\confName{CVPR}
\def\confYear{2023}

\begin{document}

\title{Rethinking Person Re-identification from a Projection-on-Prototypes Perspective}

\author{
Qizao Wang \and
Xuelin Qian \and
Bin Li \and
Yanwei Fu \and
Xiangyang Xue \and
~~~~Fudan University~~~~\and
{\tt\small qzwang22@m.fudan.edu.cn}, 
{\tt\small \{xlqian,libin,yanweifu,xyxue\}@fudan.edu.cn}
}
\maketitle

\begin{abstract}
Person Re-IDentification (Re-ID) as a retrieval task, has achieved tremendous development over the past decade. Existing state-of-the-art methods follow an analogous framework to first extract features from the input images and then categorize them with a classifier. However, since there is no identity overlap between training and testing sets, the classifier is often discarded during inference. Only the extracted features are used for person retrieval via distance metrics. In this paper, we rethink the role of the classifier in person Re-ID, and advocate a new perspective to conceive the classifier as a projection from image features to class prototypes.
These prototypes are exactly the learned parameters of the classifier. In this light, we describe the identity of input images as similarities to all prototypes, which are then utilized as more discriminative features to perform person Re-ID. We thereby propose a new baseline ProNet, which innovatively reserves the function of the classifier at the inference stage. To facilitate the learning of class prototypes, both triplet loss and identity classification loss are applied to features that undergo the projection by the classifier. An improved version of ProNet++ is presented by further incorporating multi-granularity designs. Experiments on four benchmarks demonstrate that our proposed ProNet is simple yet effective, and significantly beats previous baselines. ProNet++ also achieves competitive or even better results than transformer-based competitors.
\end{abstract}

\section{Introduction} \label{sec:intro}

\begin{figure}[t]
    \centering
    \includegraphics[scale=0.31]{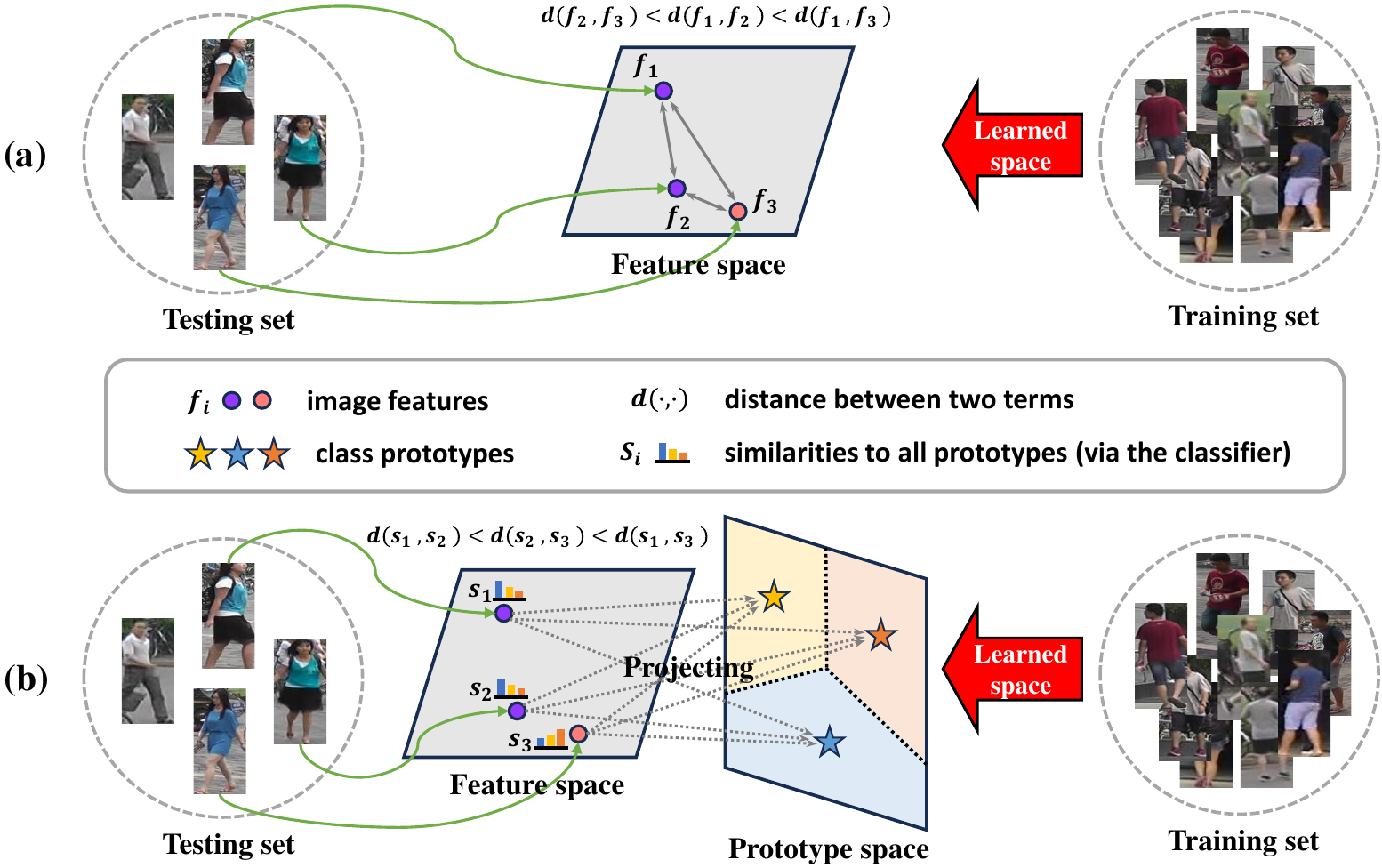}
    \caption{
    \textbf{Comparison with the conventional person Re-ID pipeline.} Once the feature space is learned from the training images, \textbf{(a)} the conventional Re-ID pipeline discards the classifier at the testing stage, and only the extracted features are used for person retrieval via distance metrics. Instead, 
    \textbf{(b)} we reserve the classifier during inference to project features onto class prototypes, \textit{i.e.}, describing the input as the similarity to all class prototypes. Re-ID is thus performed by comparing the distribution difference of similarities.
    }
    \label{fig:intro}
\end{figure}

Person Re-IDentification (Re-ID) aims to recognize the same person from multiple images captured by different cameras at different times, which has received extensive attention and research over the past decade. With the renaissance of deep convolutional networks, especially the present of residual networks, the deep learning based Re-ID approaches have showcased significant progress, even surpassing the performance of humans ~\cite{zhang2017alignedreid,luo2019alignedreid++}. 
Existing deep methods predominantly embrace an analogous framework to image classification \cite{zheng2017discriminatively, qian2019leader,li2018harmonious,ye2021deep}, which consists of a feature extractor to distill visual features from person images and a classifier to categorize these features based on identity labels. However, person Re-ID is essentially an instance-level retrieval task that is fundamentally different from the category-level classification task. 

As a retrieval task, person identities are non-overlapped during training and testing. The classifier thus becomes obsolete upon the deployment of a person Re-ID model, rendering the focus onto the feature extractor. As shown in Fig.~\ref{fig:intro} (a), the extracted features are then harnessed, employing a distance metric, \textit{e.g.}, Euclidean or cosine distance, to effectuate the retrieval of the target pedestrian. Therefore, previous efforts are predominantly channeled toward the improvement of feature extractors. For example, PCB \cite{sun2018beyond} partitions feature maps into different horizontal stripes, so as to learn stronger representations from multiple parts. Luo \textit{et al.} \cite{luo2019bag} further improve it by introducing Batch Normalization Neck (BNNeck) and bag of tricks. Zhou \textit{et al.}~\cite{zhou2019omni} design a omni-scale network to learn features of both homogeneous and heterogeneous scales. Recently, TransReID~\cite{he2021transreid} builds a pure transformer-based framework to overcome the information loss in convolutional networks. 
These are strong baselines that have witnessed the development of person Re-ID tasks.

In this paper, we rethink the role of the classifier in person Re-ID pipeline. It is originally designed to learn decision boundaries of categories so that identity labels can promote the learning of feature extractors via back-propagation. We argue that abandoning the classifier not only fails to make full use of the known identity knowledge, but also leads to gaps between the training and testing phases.
Conceptually, it is abstract to directly define the identity of a new person, humans often perform it by describing his/her dissimilarities and similarities to others. 
On the one hand, when comparing each person with others, some detailed and critical identity differences can be discovered to better distinguish between each other. On the other hand, in a comparable vein, if two individuals are described in an analogous way by others, there is a high probability that they are the same person.

Following this underlying motivation, we provide a new perspective on the person Re-ID pipeline, by conceiving the parameters of the classifier as overcomplete bases. In this light, the computational process of the classifier can be understood as projecting image features onto these bases, \textit{i.e.}, describing similarities of the input to all bases, as shown in Fig.~\ref{fig:intro} (b). We call these bases class prototypes, since they are learned from the identity/category space composed of images from the training set. Accordingly, we propose a new baseline termed ProNet, which leaves the classifier during inference and takes advantage of projected features as more discriminative representations. Technically, ProNet has a very simple architecture, where it takes ResNet-50~\cite{resnet} as the feature extractor with consideration of its popularity in the person Re-ID community, and one Fully-Connected (FC) layer as the classifier. Inspired by LDA~\cite{xanthopoulos2013linear}, we innovatively apply the triplet loss~\cite{zhai2019defense} to features that undergo the projection by the classifier, rather than to those extracted by the feature extractor as all previous methods do~\cite{sun2018beyond, luo2019bag, zhou2019omni}. Combined with identity classification loss~\cite{luo2019bag}, both of them can supervise the learning of class prototypes and ensure their effectiveness.
Experiments in Tab.~\ref{tab:ablation} strongly suggest that such simple designs can boost the performance by +2.6\% and +13.5\% mAP on the Market-1501 and CUHK03 datasets, respectively, serving as a strong baseline for Re-ID. Furthermore, we seamlessly incorporate the merits of PCB \cite{sun2018beyond} into our framework (termed ProNet++), endowing ProNet with multi-granularity information. It exhibits a more powerful ability, outperforming transformer-based methods.
\textbf{Our contributions are summarized as follows,}

\textbf{(1)} To the best of our knowledge, we are the first to advocate leaving the classifier during inference to promote person Re-ID performance. Accordingly, we provide new insights on identity classification loss and triplet loss to exert the positive impact of class prototypes for Re-ID.

\textbf{(2)} We propose a strong baseline, namely ProNet, which takes advantage of class prototypes to achieve competitive performance with recent advanced methods. To give full play to class prototypes and show the generalization, an improved version (ProNet++) taking advantage of part-level information with the proposed multi-granularity fusion module is proposed.

\textbf{(3)} Our proposed strong baseline achieves state-of-the-art performance on four Re-ID datasets without bells and whistles. Thorough experiments demonstrate the effectiveness of class prototypes during training and inference.

\section{Related Work}
\subsection{Person Re-Identification}
Recently, lots of efforts have been paid to promote person Re-ID. For instance, to reduce the influence of intra-class variations, \cite{zheng2019joint} seeks to improve learned Re-ID embeddings by better leveraging the generated data. \cite{zhou2019omni} proposes to perform omni-scale feature learning by designing a residual block composed of multiple convolutional streams, each detecting features at a certain scale.
Besides, part-level features have been proven effective for person Re-ID. Previous part-based methods focus on locating regions with prior knowledge like poses or body landmarks~\cite{su2017pose,zhao2017spindle,sarfraz2018pose}. Instead of using external cues, \cite{sun2018beyond} uses uniformly partitioned horizontal stripes to directly learn local representations. Based on the part-based convolutional baseline PCB~\cite{sun2018beyond}, \cite{zhang2017alignedreid} dynamically aligns local information of two person images with the shortest path algorithm. \cite{wang2018learning} proposes a multi-branch deep network architecture integrating discriminative information with various granularities. \cite{zheng2019pyramidal} and \cite{fu2019horizontal} explore multi-granularity feature representations at different pyramid scales. Some methods turn to auxiliary data clues~\cite{jin2020semantics,zhu2020aware} or modalities~\cite{guo2019beyond,gao2020pose} but face the inflexibility of relying on additional information. 
Previous works in person Re-ID mainly focus on improving the feature extraction capability of the model. Particularly, we notice one work \cite{yu2019unsupervised} that shares a similar insight to describe images with the similarity to class prototypes. \cite{yu2019unsupervised} studies the task of unsupervised person Re-ID, and uses additional data to learn prototypes. Differently, in this paper, we emphasize the non-negligible role of the classifier in the person Re-ID pipeline. We propose a new baseline to reuse the classifier at the testing stage, where the learned parameters of the classifier are treated as class prototypes.

\begin{figure*}
    \centering
    \includegraphics[scale=0.7]{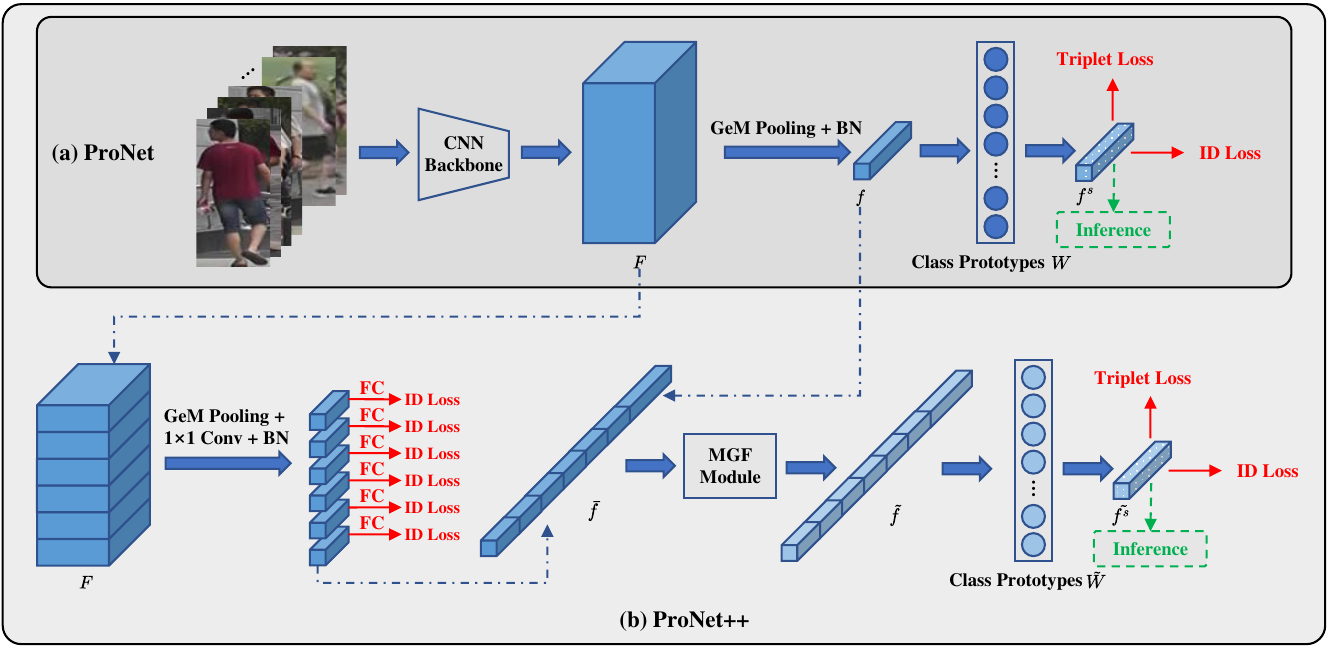}
    \caption{\textbf{Framework of (a) our proposed strong baseline ProNet and (b) the multi-granularity enhanced ProNet++.}  The losses and modules in red only participate in the training phase, and the green line indicates used features during inference. The weights of the classifier are regarded as class prototypes for effective comparison between identities. Both ID loss and triplet loss are applied to $f^{s}$, which is also used for inference, with new insights to take advantage of the positive role of class prototypes. To further improve the ability of class prototypes, multi-granularity enhanced identity feature is obtained with Multi-Granularity Fusion (MGF) module applied on the global feature and part features.}
    \label{fig:framework}
\end{figure*}

\subsection{Interacting Between Images for Re-ID}
If learning representations from single images, instance-level features can ignore the discriminative information since the appearance of each identity varies in different images. Therefore, some works explore interactions between different person images.  For example, \cite{shen2018person} models pairwise relationships between different probe-gallery pairs with deeply learned message passing on a graph. \cite{zhang2021person} proposes the heterogeneous local graph attention networks to model the intra-local relation among parts within the single pedestrian image and the inter-local relation among parts from different images. \cite{wang2022nformer} explicitly models interactions across all input images, thus suppressing outlier features and leading to more robust representations overall. 

During inference, these methods would model the relations between all the images from the query and gallery sets. It is computationally prohibitive, even if some efforts have been made to relieve the resource burden~\cite{wang2022nformer}. What's more, if the number of images for each identity in the testing set is small, individuals can obtain limited useful information from each other~\cite{wang2022nformer}. Considering the consuming extra computation cost, \cite{zhu2022dual} proposes to establish interaction between image pairs during training to regularize the attention learning of an image and remove the interaction during inference. 
Differently, we propose to leave the classifier used during training to promote inference performance. Our proposed ProNet can effectively benefit from interactions with class prototypes that are more robust than image instances. More importantly, there is rarely additional computational cost for both training and inference.

\section{Methdology}
In this section, we first introduce the overall framework of our ProNet. Then, we elaborate on the new insights of the classifier in person Re-ID and the usage of triplet loss in Sec.~\ref{subsec:cls_layer} and Sec.~\ref{subsec:tri_loss}, respectively. Finally, we further improve ProNet with multi-granularity features in Sec.~\ref{subsec:pronet++}, showing its generalization and effectiveness. Our proposed framework is shown in Fig.~\ref{fig:framework}.

\subsection{Overview of ProNet} \label{subsec:preliminary}
\noindent \textbf{Feature extraction.} 
Given a training dataset $\mathcal{D} = \{ x_{i}, y_{i}\}_{i=1}^{N}$ containing $N$ images and $N^{p}$ identities, where $x_{i}$ and $y_{i}$ denotes the $i$-$th$ images and its corresponding identity label, a CNN model $\mathcal{G}$ is applied to extract image features.
Following the common practice~\cite{wang2018learning,ye2021deep}, ResNet-50~\cite{resnet} initialized by ImageNet~\cite{deng2009imagenet} without the average pooling and the classifier is adopted as the backbone, and the stride of the first convolution layer in $res4$ block is set to 1 to increase the feature resolution. 
$P$ identities and $K$ images per person are randomly sampled to constitute a training batch. For each image $x_{i}$ in the training batch $\mathcal{B}$, feature map $F_{i} = \mathcal{G}\left(x_{i}\right) \in \mathbb{R}^{h \times w \times d}$ can be obtained, where $h$ and $w$ are the height and width of the feature map, and $d$ denotes the feature dimension.

With $F_{i}$, Generalized-Mean (GeM) pooling~\cite{radenovic2018fine} is adopted as a compromise between GAP and GMP to output the identity feature $f_{i} \in \mathbb{R}^{d}$. By convention, either Global Average Pooling (GAP) or Global Max Pooling (GMP) is usually adopted to obtain the feature vector representation for each image. On the one hand, GAP tends to produce smooth features against noises, but it can lose some discriminative information due to the average operation. For example, the body region and the background region are treated equally. On the other hand, GMP can focus on discriminative parts, but can also be fooled by some noisy information. Therefore, combining both pooling operations can help extract more robust and discriminative identity features. We serve the learnable Generalized-Mean pooling (GeM)~\cite{radenovic2018fine} as a compromise between GAP and GMP, outputting the identity feature $f_{i} \in \mathbb{R}^{d}$. Formally,
\begin{equation}
    f_{i} = {\left( \frac{1}{h \times w} \sum_{a \in A_{i}} a^{n}\right)}^{\frac{1}{n}}
\label{eq:gem_pool}
\end{equation}
where $A_{i}$ is the set of $h \times w$ activations for the feature map $F_{i}$, and $n$ is the learnable pooling parameter. Eq.~\ref{eq:gem_pool} is equivalent to GAP when $n = 1$ and GMP when $n \to \infty$. 
The effectiveness of GeM pooling is further discussed in Sec.~\ref{subsec:ablation}.

\noindent \textbf{Optimization.} 
As the common practice in person Re-ID, identity classification loss (ID loss) and triplet loss~\cite{hermans2017defense} are used jointly for optimization. Unless otherwise specified, we omit the subscript $i$ of $F_{i}$ and $f_{i}$ in the following to simplify notation, and denote the feature after BNNeck as $f$.

\subsection{Revisiting the Role of Classifier} \label{subsec:cls_layer}
\noindent \textbf{Common practice.} 
A Fully-Connected (FC) layer is appended after the BNNeck as the identity classifier to calculate ID loss. More specifically, the classifier with learnable parameters $W \in \mathbb{R}^{N^{p} \times d}$ outputs the identity class prediction, following a softmax function to calculate $\mathcal{L}_{id}$. The bias of the classifier is kept to 0 following~\cite{luo2019bag}. To prevent overfitting for a classification task, label smoothing regularization~\cite{inception_v3} is usually applied. Formally,
\begin{equation}
    \mathcal{L}_{id} = - \frac{1}{|\mathcal{B}|} \sum\limits_{f \in \mathcal{B}} \sum\limits_{j = 1}^{N^{p}} q_{j} \ \log \frac{\exp (f \cdot w_{j}^{\top})}{\sum\limits_{k=1}^{N^{p}} \exp (f \cdot w_{k}^{\top})}
\label{eq:cls_loss}
\end{equation}
\begin{equation}
    q_{j} = 
    \left\{
    \begin{aligned}
      &1 - \epsilon + \frac{\epsilon}{K} &&,j = y_{i} \\  
      &\frac{\epsilon}{K} &&, {\rm otherwise}
    \end{aligned}
    \right.
\label{eq:cls_LS}
\end{equation}
\noindent where $|\mathcal{B}| = P \times K$ denotes the training batch size, $w_{j} \in \mathbb{R}^{d}$ denotes the $j$-$th$ row of $W$, and $\epsilon$ is a small constant for label smoothing regularization, which is simply set to 0.1.

In the person Re-ID evaluation protocol, the pedestrian identity categories during training and testing do not intersect, that is, the identities in the testing set are all unseen in the training set. Therefore, \textbf{(1)} during inference, the classifier is always discarded for all existing Re-ID methods and the feature before the classifier is thus used. \textbf{(2)} Label smoothing regularization prevents the Re-ID model from overfitting training identities, improving the model performance during inference.

\noindent \textbf{New perspectives.} 
The dimension of the classifier's parameters $W$ is $N^{p} \times d$, where $N^{p}$ is the number of training identities and $d$ means the feature dimension. Inspired by \cite{liu2017sphereface,wang2018cosface,deng2019arcface}, we regard $W$ as class prototypes (or overcomplete bases). Since there is no identity overlap between the training and testing set, we agree that the classifier cannot be used to recognize the identity of testing images. However, we provide a new perspective of \textit{conceiving the computational process of the classifier as projecting image features onto these prototypes.} 
Although for identities that appear during inference, the class prototypes $W$ stand for unseen identities that appeared during training, the same identity should have consistent relative similarities with these class prototypes. Intuitively, we can describe a person as being more like someone, somewhat like someone, and less like another person, so the relative similarities can also be seen as identity features to effectively distinguish people. More concretely, we refine $f$ with the help of $W$ as follows:
\begin{equation}
    f^{s} = f \cdot W^{\top}
\label{eq:fs_compute}
\end{equation}
where $f^{s} \in \mathbb{R}^{N^{d}}$ is regarded as the identity feature and used with cosine distance metric to do the retrieval in the inference stage. Note that we do not apply the softmax function to $f^{s}$. Similarly, Eqs.~\ref{eq:cls_loss} and \ref{eq:cls_LS} can be adopted for optimization in the training stage. Differently, label smoothing regularization can also play another important role that empowering the model with the ability to compare different class prototypes. Taking negative classes into account in Eq.~\ref{eq:cls_LS}, $f^{s}$ can benefit from all class prototypes during both training and inference.

\noindent \textbf{Further discussions.}
\textbf{(1)} The class prototypes $W$ are trained together with the backbone in an end-to-end manner, so the extracted identity feature $f$ and $W$ are in the joint latent space, making them comparable with each other. Therefore, $f$ can benefit from comparison with the class prototypes $W$ to achieve a more discriminative identity feature during both training and inference.
\textbf{(2)} Each class prototype is the feature center representation of multiple pedestrian samples with the same identity, under different cameras as well as with all kinds of variations (\textit{e.g.}, posture, viewpoint, scale variations). In other words, the learned class prototypes are more robust to various noises than instances, and independent of camera changes. Benefiting from robust class prototypes, $f^{s}$ would be more robust than $f$ and can bring better performance.

\subsection{Boosting Prototypes with Triplet Loss} \label{subsec:tri_loss}
\noindent \textbf{Common practice.} 
Triplet loss is widely used to enforce a margin $m$ between the intra and inter distances to the same anchor sample, which is formulated as:
\begin{equation}
    \mathcal{L}_{tri} = [m + \mathcal{D}(f_{a}, f_{n}) - \mathcal{D}(f_{a}, f_{p})]_{+}
\label{eq:tri_loss}
\end{equation}
where $(f_{a}, f_{p}, f_{n})$ are extracted identity features of the anchor, positive and negative sample, receptively. The positive/negative sample has the same/different identity as the anchor sample. $\mathcal{D}$ denotes the distance measure. According to ~\cite{hermans2017defense} which selects the hardest positive and the hardest negative samples within the batch, with Euclidean distance as metrics can effectively promote learning.

\noindent \textbf{New perspectives.} For our proposed ProNet, as mentioned in Sec.~\ref{subsec:cls_layer}, we advocate regarding $f^{s}$ as the identity feature. The key is to learn high-quality class prototypes, so that $f^{s}$ can be trusted. Following LDA \cite{xanthopoulos2013linear}, we intend to optimize the learning of class prototypes by directly supervising the representations of $f^{s}$. Concretely, we replace $f_{(*)}$ with $f_{(*)}^{s}$ in Eq.~\ref{eq:tri_loss}. Furthermore, we change $\mathcal{D}$ from the original Euclidean distance to cosine distance, which is consistent with the distance metric in the inference stage.

\noindent \textbf{Further discussions.}
\textbf{(1)} ID loss with label smoothing regularization simply assigns a small same weight to negative classes. Therefore, on the one hand, the model learns to be aware of the relative relationships between class prototypes. On the other hand, different class prototypes $W$ are considered to have the same similarity as input feature $f$, which restricts different class prototypes from achieving varying degrees of similarity response.
However, directly applying triplet loss on $f^{s}$ would help make the identity features of different people distinguishable, reflecting the different similarity contributions of different class prototypes.
\textbf{(2)} The process of applying triplet loss on $f^{s}$ is the same as that of pedestrian retrieval using $f^{s}$. Specifically, the triplet loss optimization process during training is to compare the similarity between $f$ and all class prototypes $W$, and then identify pedestrians based on $f^{s}$. The retrieval process during inference is to compute the similarity between the query image and all gallery images using $f^{s}$, and then identify pedestrians based on the similarity distance. The consistency in the design promotes Re-ID performance significantly (refer to Tab.~\ref{tab:ablation}).

\subsection{ProNet++} \label{subsec:pronet++}
We argue that since the class prototypes $W$ are trained together with the backbone in an end-to-end manner, a better feature representation $f$ would also contribute to better class prototypes, resulting in a better identity feature $f^{s}$ after projection. To prove that, we also try taking advantage of part-level features to enhance feature representation. The improved version, \textit{i.e.}, ProNet++, would first extract global feature and part features, then fuse them to get the multi-granularity enhanced identity feature, as shown in Fig.~\ref{fig:framework}~(b).

More specifically, the global feature $f$ is obtained as described in Sec.~\ref{subsec:preliminary}. To get part features, $F$ is horizontally partitioned into $p$ stripes and each stripe can be considered as a body part. Afterward, pooling is applied to each stripe, following a $1 \times 1$ convolutional layer to reduce the dimension of each part feature to $d'$. Lastly, each part feature $f^{p}_{j}|_{j=1}^{P}$ is supervised with ID loss $\mathcal{L}_{j}^{l}$, respectively.

\noindent \textbf{Multi-granularity fusion.}
Since there may be redundancy between features of different granularities, a Multi-Granularity Fusion (MGF) module is proposed to fuse the global feature and the local features. Inspired by~\cite{hu2018squeeze}, two fully-connected layers are applied to achieve the information bottleneck effect. Formally,
\begin{equation}
    \bar{f} = [f; f^{p}_{1}; \cdots; f^{p}_{P}]
\end{equation}
\begin{equation}
    \tilde{f} = \sigma(\phi(\bar{f} W_1 + b_1)W_2 + b_2) \otimes \bar{f}
\end{equation}
\noindent where $[*~;~*]$ denotes the concatenation operation in the feature dimension, $\bar{f} \in \mathbb{R}^{D}$, $D = d + P \times d'$. $\phi$ denotes the \textit{ReLU} activation function, $\sigma$ denotes the \textit{Sigmoid} function, and $\otimes$ denotes element-wise multiplication. $W_1 \in \mathbb{R}^{D \times (D/r)}$, $W_2 \in \mathbb{R}^{(D/r) \times D}$, $b_1 \in \mathbb{R}^{D/r}$, and $b_2 \in \mathbb{R}^{D}$ are the weights and biases of two fully-connected layers.

\begin{table*}[t]
  \centering
  \caption{\textbf{Comparison with the state-of-the-art methods on Market-1501, CUHK03 and MSMT17.} ``MG'' denotes methods using multi-granularity features. ``\#Params'' denotes the number of model parameters, shown by the relative value of comparing each method to ProNet. Methods in the gray region adopt Transformer, and methods marked with ``$*$'' use extra data or annotations. The best results of our proposed methods and comparison methods are shown in bold and underlined, respectively.}
  \label{tab:compare_SOTA}
    \setlength{\tabcolsep}{2mm}{
    \begin{tabular}{c|c|l|cc|cc|cc|cc}
    \Xhline{0.8pt}
    \multirow{2}[0]{*}{\bf Methods} & \multirow{2}[0]{*}{\bf MG} & \multirow{2}[0]{*}{\bf \#Params} & \multicolumn{2}{c|}{\bf Market-1501}  & \multicolumn{2}{c|}{\bf CUHK03(L)} & \multicolumn{2}{c|}{\bf CUHK03(D)} & \multicolumn{2}{c}{\bf MSMT17} \\
    \cline{4-11}
    \multicolumn{1}{c|}{} & \multicolumn{1}{c|}{} & \multicolumn{1}{c|}{} & mAP & Rank-1 & mAP & Rank-1 & mAP & Rank-1 & mAP & Rank-1 \\
    \Xhline{0.5pt}

    \rowcolor{gray!20} \multicolumn{1}{l|}{TransReID$^{*}$ \cite{he2021transreid}} & $\checkmark$ & 4.14x & 88.2 & 95.0 & - & - & - & - & 64.9 & 83.3 \\
    \rowcolor{gray!20} \multicolumn{1}{l|}{DCAL$^{*}$ \cite{zhu2022dual}} &  & $>$ 4.14x  & 87.5 & 94.7 & - & - & - & - & 64.0 & 83.1 \\
    \rowcolor{gray!20} \multicolumn{1}{l|}{NFormer \cite{wang2022nformer}} & & 1.10x & 91.1 & 94.7 & 78.0 & 77.2 & 74.7 & 77.3 & 59.8 & 77.3 \\
    \Xhline{0.5pt}

    \multicolumn{1}{l|}{PCB~\cite{sun2018beyond}} & $\checkmark$ & 1.01x & 77.3 & 92.4 & - & - & 54.2 & 61.3 & - & - \\
    \multicolumn{1}{l|}{PCB+RPP~\cite{sun2018beyond}} & $\checkmark$ & $>$ 1.01x & 81.6 & 93.8 & - & - & 57.5 & 63.7 & - & - \\
    \multicolumn{1}{l|}{MGN~\cite{wang2018learning}} & $\checkmark$ & 2.81x & 86.9 & 95.7 & 67.4 & 68.0 & 66.0 & 66.8 & - & - \\
    \multicolumn{1}{l|}{BOT~\cite{luo2019bag}} & & 1x & 85.9 & 94.5 & 61.3 & 62.6 & 58.4 & 60.5 & - & - \\ 
     \multicolumn{1}{l|}{Pyramid~\cite{zheng2019pyramidal}} & $\checkmark$ & 2.19x & 88.2 & 95.7 & 76.9 & 78.9 & \underline{74.8} & 78.9 & - & - \\
    \multicolumn{1}{l|}{OSNet~\cite{zhou2019omni}} & & 0.10x & 84.9 & 94.8 & - & - & 67.8 & 72.3 & 52.9 & 78.7 \\
    \multicolumn{1}{l|}{DG-Net~\cite{zheng2019joint}} & & 2.71x & 86.0 & 94.8 & - & - & - & - & 52.3 & 77.2 \\
    \multicolumn{1}{l|}{ABD-Net~\cite{chen2019abd}} & & 2.76x & 88.3 & 95.6 & - & - & - & - & \underline{60.8} & \underline{82.3} \\
    \multicolumn{1}{l|}{HPM~\cite{fu2019horizontal}} & $\checkmark$ & 1.45x & 82.7 & 94.2 & - & - & 57.5 & 63.9 & - & - \\
    \multicolumn{1}{l|}{SAN$^{*}$~\cite{jin2020semantics}} & & 1.36x & 88.0 & \underline{96.1} & 76.4 & 80.1 & 74.6 & 79.4 & 55.7 & 79.2 \\
    \multicolumn{1}{l|}{Relation-Net~\cite{park2020relation}} & $\checkmark$ & 1.99x & \underline{88.9} & 95.2 & 75.6 & 77.9 & 69.6 & 74.4 & - & - \\
    \multicolumn{1}{l|}{RGA-SC~\cite{zhang2020relation}} & & 1.25x & 88.4 & \underline{96.1} & \underline{77.4} & \underline{81.1} & 74.5 & \underline{79.6} & 57.5 & 80.3 \\
    \multicolumn{1}{l|}{CDNet~\cite{li2021combined}} & & $>$ 0.07x & 86.0 & 95.1 & - & - & - & - & 54.7 & 78.9 \\ 
    \multicolumn{1}{l|}{CAL~\cite{rao2021counterfactual}} & & 2.51x & 87.0 & 94.5 & - & - & - & - & 56.2 & 79.5 \\

    \Xhline{0.5pt}
    \multicolumn{1}{l|}{ProNet} & & 1x & 89.2 & 95.6 & 80.0 & 82.7 & 76.7 & 80.1 & 61.3 & 82.9 \\
    \multicolumn{1}{l|}{ProNet++} & $\checkmark$ & 1.43x & \textbf{90.2} & \textbf{96.0} & \textbf{82.7} & \textbf{85.2} & \textbf{79.1} & \textbf{82.6} & \textbf{65.5} & \textbf{85.4} \\ \hline
    \multicolumn{1}{l|}{ProNet (RK~\cite{zhong2017re})} & & 1x & 94.8 & 96.1 & 90.1 & 88.7 & 88.0 & 86.6 & 77.6 & 86.9 \\
    \multicolumn{1}{l|}{ProNet++ (RK~\cite{zhong2017re})} & $\checkmark$ & 1.43x & \textbf{95.3} & \textbf{96.4} & \textbf{91.9} & \textbf{90.6} & \textbf{89.2} & \textbf{87.8} & \textbf{80.0} & \textbf{88.2} \\
    \Xhline{0.8pt}
    \end{tabular}}
\end{table*}

\noindent \textbf{Training and inference.}
With the fused multi-granularity feature $\tilde{f}$, an FC layer is adopted as the identity classifier, whose weights are served as multi-granularity class prototypes $\tilde{W}$. Subsequently, $\tilde{f^{s}} = \tilde{f} \cdot \tilde{W}^{\top}$ can be obtained to calculate ID loss $\mathcal{L}_{id}^{m}$ and triplet loss $\mathcal{L}_{tri}^{m}$ in the similar way as $\mathcal{L}_{id}$ and $\mathcal{L}_{tri}$. The overall loss of ProNet++ can be formulated as:
\begin{equation}
     \mathcal{L} = \mathcal{L}_{id} + \mathcal{L}_{tri} + \frac{1}{P} \sum\limits_{j = 1}^{P} \mathcal{L}_{j}^{l} + \mathcal{L}_{id}^{m} + \mathcal{L}_{tri}^{m}
\end{equation}
Similar to our proposed ProNet, the feature after the classifier is regarded as identity representation. In other words, ProNet++ uses $\tilde{f^{s}}$ with multi-granularity identity information for inference.


\section{Experiments}

\subsection{Experimental Setup}
\noindent \textbf{Datasets.}
We evaluate our methods on three widely-used person ReID datasets, \textit{i.e.}, Market-1501 \cite{market1501}, CUHK03~\cite{li2014deepreid}, and the large-scale dataset MSMT17~\cite{wei2018person}.
\textbf{Market-1501} is captured by 6 cameras at Tsinghua University, containing 12,936 images of 751 identities for training and 19,281 images of 750 identities for testing.
\textbf{CUHK03} consists of 1,467 pedestrians. There are both manually labeled bounding boxes from 14,096 images and DPM-detected bounding boxes from 14,097 images. Following \cite{zhong2017re}, 767 identities are used for training and 700 identities for testing. We denote the labeled and detected versions as ``(L)'' and ``(D)'', respectively.
\textbf{MSMT17} is a challenging large-scale benchmark, containing 126,441 images captured by 15 cameras, where 1,041 identities and 3,060 identities are used for training and testing, respectively. 

\noindent \textbf{Evaluation metrics.} We follow the common practices to adopt Cumulative Matching Characteristics (CMC) at Rank-1 and mean Average Precision (mAP) for evaluating the performance of person Re-ID models. Of the two, mAP evaluates multiple retrieval results more comprehensively.

\noindent \textbf{Implementation details.}
Our method is implemented on PyTorch. Following~\cite{luo2019bag}, random horizontal flipping, padding, random cropping, and random erasing~\cite{zhong2020random} are used for data augmentation. The input images are resized to $384 \times 192$. The batch size is 64 with $P=8$ and $K=8$. Adam optimizer~\cite{kingma2014adam} with weight decay of $5 \times 10^{-4}$ is adopted, with the warmup strategy that linearly increases the learning rate from $3.5 \times 10^{-5}$ to $3.5 \times 10^{-4}$ in the first 10 epochs. Then, the learning rate is decreased by a factor of 10 at the 30-$th$ and the 60-$th$ epoch, respectively. The margin $m$ is set to 0.3. For ProNet++, $P$ is set to 8. Following~\cite{sun2018beyond,fu2019horizontal}, $d'$ is empirically set to 256. Following~\cite{hu2018squeeze}, $r$ is set to 16.

\subsection{Comparison with State-of-the-Art Methods}
Tab.~\ref{tab:compare_SOTA} shows the comparison results with state-of-the-art methods on three person Re-ID datasets. As introduced in Sec.~\ref{sec:intro}, BOT \cite{luo2019bag} and PCB~\cite{sun2018beyond} are two widely-used baselines. The former uses BNNeck to collaborate ID loss and triplet loss jointly, and adopt center loss~\cite{wen2016discriminative} to improve intra-class compactness. The latter takes advantage of part features for fine-grained identity information.
Compared with them, our proposed strong baseline ProNet shows much better performance with the help of class prototypes. Most methods in Tab.~\ref{tab:compare_SOTA} follow the two baselines to design more delicate feature extractors. Without bells and whistles, ProNet achieves competitive performance with advanced works in recent years.

Among all the methods, MGN~\cite{wang2018learning}, Pyramid~\cite{zheng2019pyramidal}, HPM~\cite{fu2019horizontal}, Relation-Net~\cite{park2020relation} improve PCB~\cite{sun2018beyond} for better extracting and using multi-granularity features. SAN~\cite{jin2020semantics} relies on an extra synthesized dataset. Compared with them, ProNet++ is both lightweight and effective, indicating the great effect of class prototypes.
Given the query image, RK \cite{zhong2017re} also casts light on the importance of the similarity from the comparison between pedestrians, which can further improve the performance of our proposed method.

We also compare with Transformer~\cite{vaswani2017attention} based methods, which have much more burden model parameters as shown in Tab.~\ref{tab:compare_SOTA}. Among them, both TransReID~\cite{he2021transreid} and DCAL~\cite{zhu2022dual} adopt Transformer as the backbone and use extra data for pertaining. Moreover, TransReID uses extra camera labels and DCAL takes advantage of pairwise interactions between image pairs, respectively. NFormer~\cite{wang2022nformer} uses ResNet-50 as the backbone and further adopts Transformer to model interactions across all input images, which increases computing and storage overhead during inference.
However, our proposed method is efficient during both training and inference, and shows superiority over them without relying on auxiliary clues from extra data.

\begin{table}[t]
\centering
\caption{\label{tab:vehicle_reid}\textbf{Comparasion results of our proposed ProNet and ProNet++ on the vehicle Re-ID dataset VeRi-776.} Methods in the gray region adopt Transformer, and methods marked with ``$*$'' use extra data or annotations. }
\setlength{\tabcolsep}{4mm}{
\begin{tabular}{l|cc}
\Xhline{0.8pt}
    \multicolumn{1}{c|}{\bf Methods} & \bf mAP & \bf Rank-1 \\ 
    \Xhline{0.5pt}
    \rowcolor{gray!20} \multicolumn{1}{l|}{TransReID$^{*}$ \cite{he2021transreid}} & 81.2 & 96.8 \\
    \rowcolor{gray!20} \multicolumn{1}{l|}{DCAL$^{*}$ \cite{zhu2022dual}} & 80.2 & 96.9 \\ \hline

    SPAN$^{*}$~\cite{chen2020orientation} & 68.9 & 94.0 \\
    UMTS~\cite{jin2020uncertainty} & 75.9 & 95.8 \\
    PGAN~\cite{zhang2020part} & 79.3 & 96.5 \\ 
    PVEN$^{*}$~\cite{meng2020parsing} & 79.5 & 95.6 \\
    SAVER~\cite{khorramshahi2020devil} & 79.6 & 96.4 \\
    CAL~\cite{rao2021counterfactual} & 74.3 & 95.4 \\
    GB+GFB+SLB~\cite{li2021self} & 81.0 & 96.7 \\ 
    UFDN~\cite{qian2022unstructured} & 81.5 & 96.4 \\ \hline
    
    ProNet & 81.0 & 96.8 \\
    ProNet++ & \textbf{83.4} & \textbf{97.7} \\
\Xhline{0.8pt}
\end{tabular}}
\end{table}

\noindent \textbf{Generalization to object re-identification.}
Our proposed ProNet can well generalize to other object re-identification tasks, where the class prototypes can also be obtained and utilized to promote performance. We also explore its potential for vehicle re-identification. Following existing works~\cite{meng2020parsing,khorramshahi2020devil,rao2021counterfactual,li2021self}, input images are resized to $256 \times 256$, and results on the VeRi-776~\cite{liu2016Adeep} dataset are shown in Tab.~\ref{tab:vehicle_reid}. 
Among the advanced methods, SPAN~\cite{chen2020orientation} and PVEN~\cite{meng2020parsing} take advantage of viewpoint labels. Both TransReID~\cite{he2021transreid} and DCAL~\cite{zhu2022dual} adopt ViT-B/16~\cite{dosovitskiy2020image} pre-trained on ImageNet21K and TransReID also uses both camera labels and viewpoint labels. However, our proposed ProNet achieves competitive performance without cumbersome architecture, auxiliary data, or annotation dependency. 
The improved version of ProNet++ takes advantage of the prior of local structure of human bodies. Though it is designed for person Re-ID, ProNet++ can also improve the performance of vehicle Re-ID significantly, achieving state-of-the-art performance. It is expected that with advanced designs specific to vehicles, much better results can be achieved.

\subsection{Ablation Studies} \label{subsec:ablation}

\newcommand \down[1]{ \scriptsize $\downarrow$#1}

\begin{table}[t] \footnotesize
\centering
\caption{\label{tab:ablation}\textbf{Ablation studies of ProNet and ProNet++ on Market-1501 and CUHK03(L).} ``LS'' and ``Tri.'' denote the label smoothing regularization and triplet loss, respectively. ``Infer ($\cdot$)'' denotes which feature is used for similarity measure during inference.}
\setlength{\tabcolsep}{1.1mm}{
\begin{tabular}{ll|ll|ll}
\Xhline{0.8pt}
\multicolumn{2}{c|}{\multirow{2}{*}{\bf Methods}} & \multicolumn{2}{c|}{\bf Market-1501} & \multicolumn{2}{c}{\bf CUHK03(L)} \\ 
    \cline{3-6}
    & & mAP & Rank-1 & mAP & Rank-1 \\
    \Xhline{0.5pt}
    \multirow{4}{*}{ProNet} &  \textit{w/o} LS & 87.1\down{2.1} & 94.3\down{1.3} & 77.6\down{2.4} & 79.9\down{2.8} \\
                            &  \textit{w/o} Tri. & 85.8\down{3.4} & 94.5\down{1.1} & 64.3\down{15.7} & 66.1\down{16.6} \\
                            &  \cellcolor[HTML]{defcde}Infer $f$ & \cellcolor[HTML]{defcde}86.6\down{2.6} & \cellcolor[HTML]{defcde}94.3\down{1.3} & \cellcolor[HTML]{defcde}66.5\down{13.5} & \cellcolor[HTML]{defcde}66.9\down{15.8} \\
                            &  \cellcolor[HTML]{defcde}Infer $f^{s}$ & \cellcolor[HTML]{defcde}\textbf{89.2} & \cellcolor[HTML]{defcde}\textbf{95.6} & \cellcolor[HTML]{defcde}\textbf{80.0} & \cellcolor[HTML]{defcde}\textbf{82.7} \\ \hline
    \multirow{3}{*}{ProNet++} &  \textit{w/o} MGF & 89.7\down{0.5} & 95.9\down{0.1} & 82.3\down{0.4} & 84.6\down{0.6} \\
                              &  \cellcolor[HTML]{defcde}Infer $\tilde{f}$ & \cellcolor[HTML]{defcde}88.3\down{1.9} & \cellcolor[HTML]{defcde}95.4\down{0.6} & \cellcolor[HTML]{defcde}76.1\down{6.6} & \cellcolor[HTML]{defcde}77.3\down{7.9} \\
                              &  \cellcolor[HTML]{defcde}Infer $\tilde{f^{s}}$ & \cellcolor[HTML]{defcde}\textbf{90.2} & \cellcolor[HTML]{defcde}\textbf{96.0} & \cellcolor[HTML]{defcde}\textbf{82.7} & \cellcolor[HTML]{defcde}\textbf{85.2} \\
\Xhline{0.8pt}
\end{tabular}}
\end{table}

In this subsection, we perform ablation studies of each module and hyper-parameters to demonstrate the effectiveness of class prototypes.

\noindent \textbf{Influence of label smoothing regularization.} As the results of ``ProNet \textit{w/o} LS'' in Tab.~\ref{tab:ablation} shown, there is a huge performance decrease when label smoothing regularization is removed. It confirms our statement in Sec.~\ref{subsec:cls_layer} that label smoothing regularization enables $f$ to benefit from comparison with the class prototypes $W$ to achieve a more discriminative identity feature during both training and inference.

\noindent \textbf{Effectiveness of triplet loss.} As the results of ``ProNet \textit{w/o} Tri.'' in Tab.~\ref{tab:ablation} shown, triplet loss would bring a huge performance improvement. It confirms our statement in Sec.~\ref{subsec:tri_loss} that applying triplet loss directly on $f^{s}$ helps make the identity features of different people distinguishable, reflecting the different similarity contributions of different class prototypes. Also, the design consistency between the optimization process and the retrieval process promotes Re-ID performance dramatically.

\noindent \textbf{Effectiveness of the MGF module.} For ProNet++, we introduce the MGF module to effectively fuse the global feature and the local features. The performance improvement on both datasets demonstrates the bottleneck benefits when comparing ``Ours'' with ``ProNet++ \textit{w/o} MGF'' in Tab.~\ref{tab:ablation}.
What's more, inferring with $\tilde{f^{s}}$ performs better than $f^{s}$, since $\tilde{f^{s}}$ benefits from multi-granularity enhanced class prototypes, showing the effectiveness of the design.

\begin{figure*} [t]
    \centering
    \includegraphics[scale=0.59]{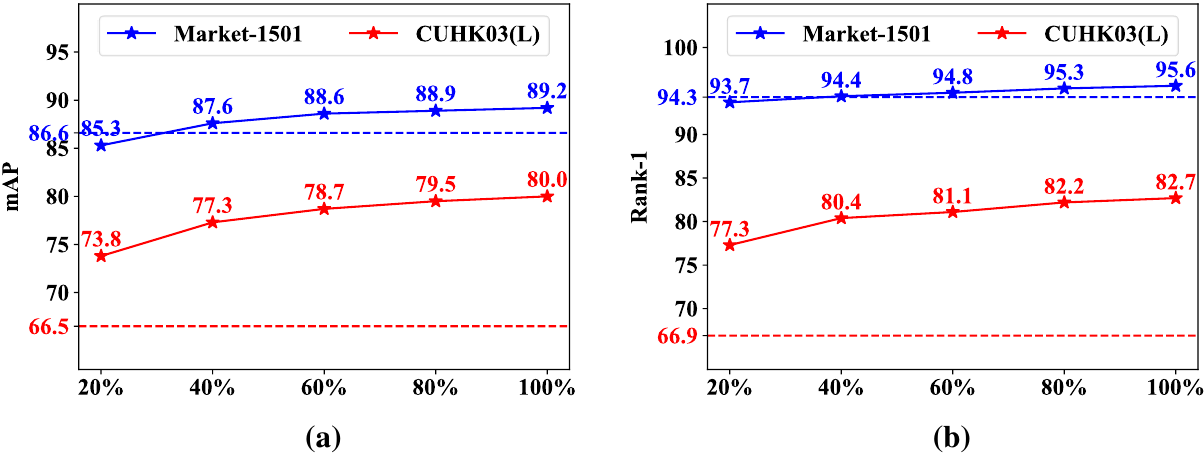}
    \caption{\textbf{Influence of the number of class prototypes for ProNet on Market-1501 and CUHK03(L).} After training, the class prototypes $W \in \mathbb{R}^{N^{d}}$ can be obtained, and ``$x \%$'' of these class prototypes are used in the testing stage, that is, $f^{s} \in \mathbb{R}^{N^{d} \times x \%}$ is used for inference. The dashed lines denote the results using the feature before the classifier (\textit{i.e.}, $f$) for inference.}
    \label{fig:ablation_num_prototypes}
\end{figure*}

\noindent \textbf{Effectiveness of class prototypes.}
As shown in the blue region of Tab.~\ref{tab:ablation}, for ProNet, inferring with $f^{s}$ shows much better performance than $f$ on both datasets, demonstrating that the learned class prototypes $W$ can make a difference in identifying and distinguishing pedestrians during inference. Similarly, with the help of class prototypes, $\tilde{f^{s}}$ shows superiority over $\tilde{f}$ on both datasets. We are the first work to bring to light the feature after the classifier for Re-ID, both in the training and the inference stage.

\noindent \textbf{Influence of the number of class prototypes.} 
Different class prototypes can generate varying degrees of similarity responses to the input identity feature $f$. As shown in Fig.~\ref{fig:ablation_num_prototypes}, as the number of class prototypes increases, better retrieval results are achieved. More class prototypes can more comprehensively represent each identity and reflect the differences between people, resulting in more discriminative identity features.
As class prototypes can be considered as overcomplete bases, fewer of them can also help achieve competitive results. Note that since the Market-1501 dataset has been extensively studied and the performance has been saturated, it appears that the accuracy margin on it is not significant. However, the significant performance advantages of ProNet on other datasets can fully demonstrate the effectiveness of it. As observed in Fig.~\ref{fig:ablation_num_prototypes}, when only 20\% class prototypes are used, $f^{s}$ is still superior to $f$ by a large margin on CUHK03 (L).

\begin{table}[t]
\centering
\caption{\label{tab:pool_ablation}Influence of different pooling strategies for ProNet on the Market-1501 and CUHK03(L) datasets.}
\setlength{\tabcolsep}{1mm}{
\begin{tabular}{l|ll|ll}
\Xhline{0.8pt}
\multicolumn{1}{c|}{\multirow{2}{*}{\bf Methods}} & \multicolumn{2}{c|}{\bf Market-1501} & \multicolumn{2}{c}{\bf CUHK03(L)} \\ 
    \cline{2-5}
        & mAP & Rank-1 & mAP & Rank-1 \\ \hline
    \multicolumn{1}{l|}{PCB~\cite{sun2018beyond}} & 77.3 & 92.4 & - & - \\
    \multicolumn{1}{l|}{PCB+RPP~\cite{sun2018beyond}} & 81.6 & 93.8 & - & - \\
    \multicolumn{1}{l|}{BOT~\cite{luo2019bag}} & 85.9 & 94.5 & 61.3 & 62.6 \\ \hline
    \multicolumn{1}{l|}{MGN~\cite{wang2018learning}} & 86.9 & 95.7 & 67.4 & 68.0 \\
    \multicolumn{1}{l|}{DG-Net~\cite{zheng2019joint}} & 86.0 & 94.8 & - & - \\ \hline

    ProNet \textit{w/} GAP & 87.5\down{1.7} & 94.9\down{0.7} & 76.5\down{3.5} & 78.9\down{3.8} \\
    ProNet \textit{w/} GMP & 88.1\down{0.5} & 95.1\down{0.5} & 78.7\down{1.3} & 81.1\down{1.6} \\
    ProNet \textit{w/} GeM & \textbf{89.2} & \textbf{95.6} & \textbf{80.0} & \textbf{82.7} \\
\Xhline{0.8pt}
\end{tabular}}
\end{table}

\noindent \textbf{Influence of different pooling strategies.} 
For previous baselines, both PCB~\cite{sun2018beyond} and BOT~\cite{luo2019bag} adopt GAP for both part and global features. Some advanced methods adopt GMP for multi-granularity features~\cite{wang2018learning,zheng2019joint}. As shown in Tab.~\ref{tab:ablation}, for our proposed ProNet baseline, GMP shows superiority over GAP, and applying GeM achieves the best results. Since GeM pooling draws on the advantages of both GAP and GMP, it is better at preserving useful identity features. Note that with different pooling strategies, our proposed ProNet baseline all performs much better than the two previous widely-used baselines PCB~\cite{sun2018beyond} and BOT~\cite{luo2019bag}.

\begin{figure} [t]
    \centering
    \includegraphics[scale=0.4]{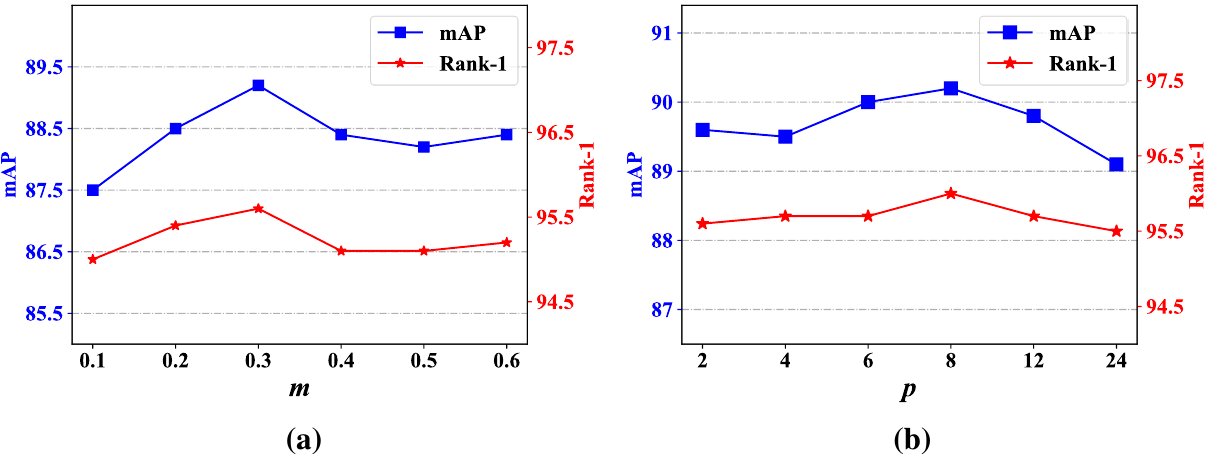}
    \caption{Ablation Studies of (a) margin $m$ of triplet loss in ProNet, and (b) the number of parts $p$ in ProNet++ on the Market-1501 dataset.}
    \label{fig:hyperparam}
    \vspace{-0.1in}
\end{figure}

\begin{figure*} [t]
    \centering
    \includegraphics[scale=0.39]{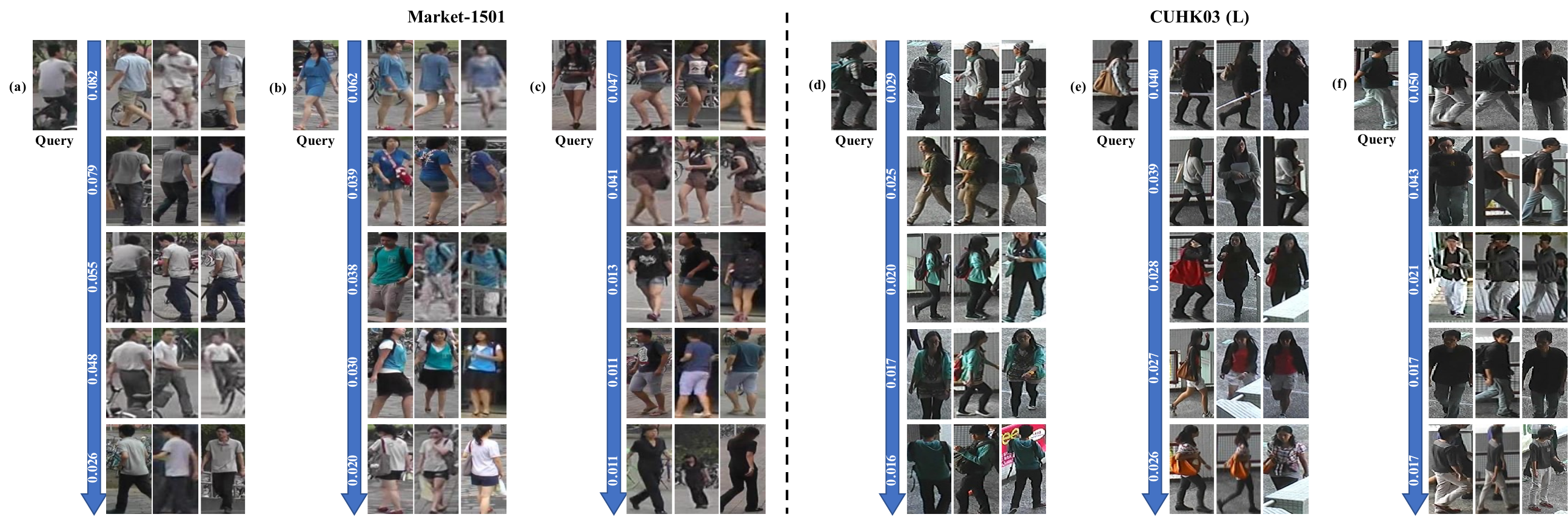}
    \caption{\textbf{Visualizations of images corresponding to class prototypes with the top five similarities with each query image during inference on Market-1501 and CUHK03 (L) datasets, respectively.} Similarity decreases from top to bottom, and the similarity scores are computed via applying the $softmax$ function on $f^{s}$. Three training images corresponding to the same identity are randomly sampled for each class prototype. Best viewed in color and zoom in.}
    \label{fig:top_prototypes}
    \vspace{-0.1in}
\end{figure*}

\noindent \textbf{Influence of margin.}
Batch hard triplet loss with margin $m$ is used to effectively promote learning in our proposed ProNet. As shown in Fig.~\ref{fig:hyperparam}~(a), $m=0.3$ achieves the best performance on Market-1501. We simply set $m=0.3$ for other datasets, except for the small-scale dataset CUHK03, we use all mined hard triplets for optimization.

\noindent \textbf{Influence of the number of parts.}
The number of parts $p$ determines the granularity of part features. As shown in Fig.~\ref{fig:hyperparam}~(b), as $p$ increases, the performance on Market-1501 increases at first. However, when $p$ is too large, each small part becomes less discriminative, containing more noise, so the model performs inferior. 
It is worth mentioning that for the baseline PCB~\cite{sun2018beyond}, which shares a similar part learning design with ProNet++. A similar trend of performance along with different numbers of parts is also observed in PCB. Different from it, we further utilize the part features for multi-granularity fusion, resulting in more robust identity features. According to the results in Fig.~\ref{fig:hyperparam}~(b), for a tradeoff between performance and efficiency, we set $p=8$ for all datasets without grid tuning.

\begin{table}[t]
\centering
\caption{\label{tab:ablation_img_size}Influence of different input image sizes on the Market-1501 and CUHK03(L) datasets. Two widely-used baselines PCB and BOT are also reproduced for comparison.}
\setlength{\tabcolsep}{1.4mm}{
\begin{tabular}{ll|cc|cc}
\Xhline{0.8pt}
\multicolumn{2}{c|}{\multirow{2}{*}{\bf Methods}} & \multicolumn{2}{c|}{\bf Market-1501} & \multicolumn{2}{c}{\bf CUHK03(L)} \\ 
    \cline{3-6}
    & & mAP & Rank-1 & mAP & Rank-1 \\ 
    \Xhline{0.5pt}
    \multirow{2}{*}{PCB~\cite{sun2018beyond}} & 256 $\times$ 128 & 73.1 & 89.2 & 52.3 & 55.1 \\
                            & 384 $\times$ 192 & 74.4 & 91.1 & 54.7 & 57.6 \\ \hline
    \multirow{2}{*}{BOT~\cite{luo2019bag}} & 256 $\times$ 128 & 85.6 & 93.8 & 61.3 & 62.6 \\
                            & 384 $\times$ 192 & 85.6 & 94.0 & 63.6 & 64.9 \\ \hline
    \multirow{4}{*}{ProNet} & 256 $\times$ 128 & 88.7 & 95.2 & 77.0 & 80.9 \\
                            & 256 $\times$ 192 & 88.6 & 95.1 & 78.5 & 81.6 \\
                            & 384 $\times$ 128 & 88.5 & 95.4 & 78.1 & 81.5 \\
                            & 384 $\times$ 192 & \textbf{89.2} & \textbf{95.6} & \textbf{80.0} & \textbf{82.7} \\ \hline
    \multirow{4}{*}{ProNet++} & 256 $\times$ 128 & 89.3 & 95.4 & 79.7 & 82.3 \\
                              & 256 $\times$ 192 & 89.5 & 95.6 & 80.6 & 83.0 \\
                              & 384 $\times$ 128 & 89.3 & 95.5 & 81.4 & 83.9 \\
                              & 384 $\times$ 192 & \textbf{90.2} & \textbf{96.0} & \textbf{82.7} & \textbf{85.2} \\
\Xhline{0.8pt}
\end{tabular}}
\end{table}

\noindent \textbf{Influence of image size.}
Various image sizes are used for comparison methods in Tab.~\ref{tab:compare_SOTA}. Three kinds of configuration are used in different methods, \textit{i.e.}, $256 \times 128$, $384 \times 128$, and $384 \times 192$. We also explore the influence of different image sizes for our proposed ProNet as well as ProNet++. Results in Tab.~\ref{tab:ablation_img_size} show that a larger image size usually brings better performance for our baselines. 
Note that with $256 \times 128$, our proposed baseline still achieves competitive performance with advanced methods (shown in Tab.~\ref{tab:compare_SOTA}), which use complex modules or architectures, or rely on auxiliary information. ProNet with $256 \times 128$ shows much better performance than past baselines with an even larger image size $384 \times 192$, demonstrating that it is indeed the class prototypes that promote Re-ID performance for our proposed ProNet.

\begin{figure*} [t]
    \centering
    \includegraphics[scale=0.45]{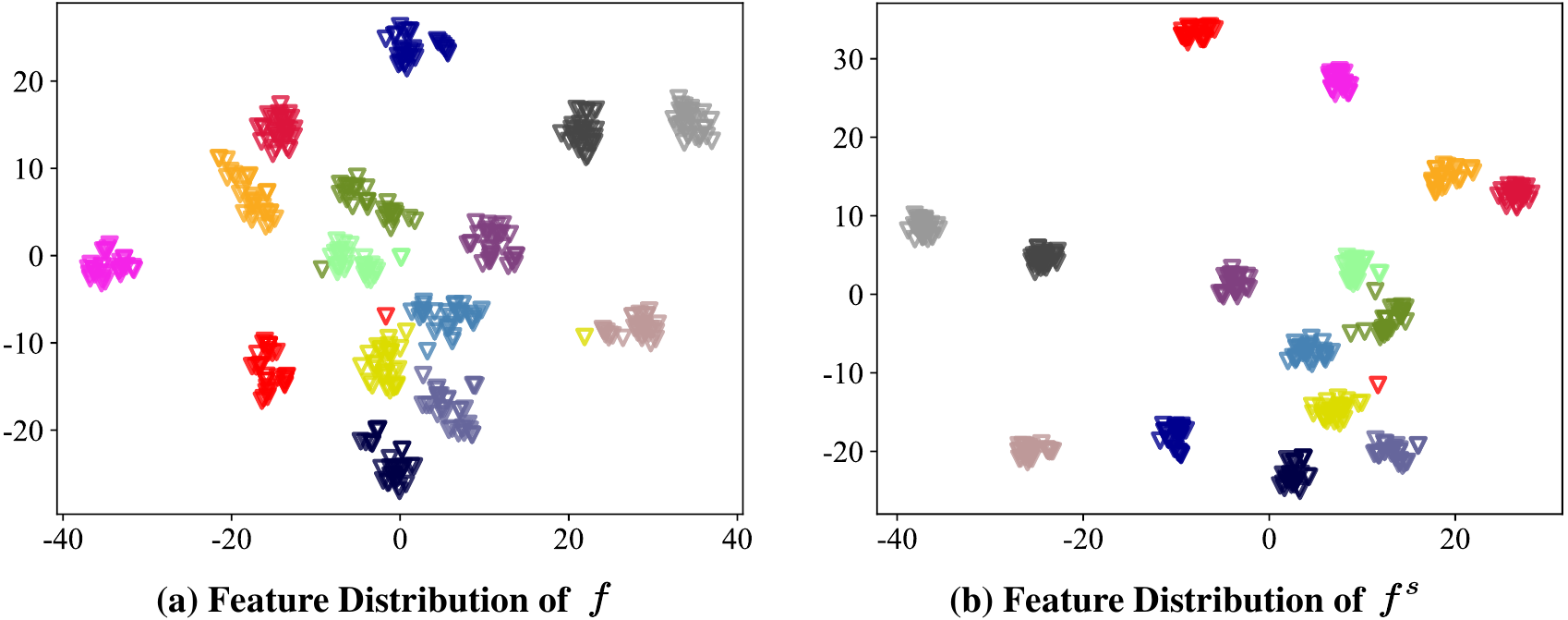}
    \caption{\textbf{t-SNE visualization of the feature distributions of $f$ and $f^{s}$ in ProNet, respectively.} Fifteen pedestrians are randomly sampled from the testing set of MSMT17, and each color represents an identity. Best viewed in color and zoom in.}
    \label{fig:feat_distribution}
\end{figure*}

\subsection{Visualizations}
In this subsection, visualizations of class prototypes, feature distributions and retrieval results are provided to better understand how class prototypes help Re-ID.

\noindent \textbf{Visualization of class prototypes.}
To better demonstrate how class prototypes work intuitively, we visualize the training images corresponding to class prototypes with the top five similarities with each query image in Fig.~\ref{fig:top_prototypes}. Note that for each query image that appeared during inference, the class prototypes correspond to unseen identities that appeared during training. 
\textbf{(1) The identities corresponding to the top similar class prototypes share similar characteristics.} For example, as shown in Fig.~\ref{fig:top_prototypes}~(a), the top similar identities all wear similar clothes, and in Fig.~\ref{fig:top_prototypes}~(e), the top similar identities all carry similar shoulder bags.
Similar body shapes can be discovered in the last row of Fig.~\ref{fig:top_prototypes}~(b) and (c). Most of the top identities share the same gender and some identities with similar characteristics but different gender can also make a difference as shown in the first and last row of Fig.~\ref{fig:top_prototypes}~(d), where similar knapsack and clothing can be seen with the query image.
\textbf{(2) As the feature center representation of multiple pedestrian samples with the same identity, the learned class prototypes are more robust to noises and variations than instances.}
Although there are great posture, viewpoint, and scale variations within different images of the same identity, the class prototypes can learn the dominant and common identity characteristics. The identities with top similarities share identity characteristics independent of variations. Class prototypes are also robust to noises like the object or pedestrian occlusions as shown in Fig.~\ref{fig:top_prototypes}~(e) and (f). Moreover, class prototypes are learned with cross-camera training samples, so they are robust to camera changes, contributing to cross-camera retrieval during inference. Benefiting from their robust representations, features after the classifier would bring better performance.

\noindent \textbf{Visualization of feature distributions.} 
To show the superiority of $f^{s}$ over $f$ intuitively, we use t-SNE~\cite{van2008visualizing} to visualize the feature distributions of them, respectively. From Fig.~\ref{fig:feat_distribution}, we can observe that
\textbf{(1)} before projecting image features to the space of class prototypes, different identities are poorly distinguished in the original image feature space, which would cause lots of false pedestrians to be retrieved. 
\textbf{(2)} The distribution of $f^{s}$ exhibits better intra-class compactness and inter-class discreteness than that of $f$. It shows performing Re-ID in the space of prototypes can be a better alternative than in the original image space. 
\textbf{(3)} Benefiting from comparison with class prototypes to discover detailed and critical identity differences, some hard samples can be retrieved correctly. For instance, one sample with the identity of $\textcolor[RGB]{227,227,51}{\nabla}$ (in yellow) is wrongly aggregated with the identity of $\textcolor[RGB]{203,173,173}{\nabla}$ (in coral red) in Fig.~\ref{fig:feat_distribution} (a), but it is corrected in Fig.~\ref{fig:feat_distribution}~(b).

\begin{figure}[t]
    \centering
    \includegraphics[scale=0.84]{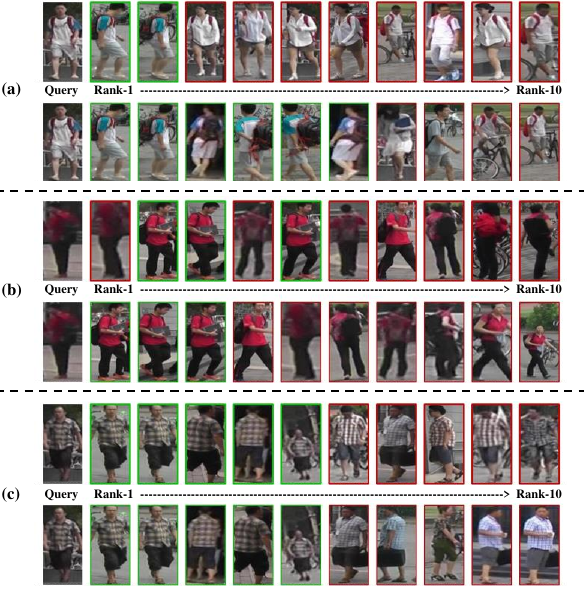}
    \caption{\textbf{Visualization of top-10 retrieval results on the Market-1501 dataset.} For each query image, the first and the second row are the ordered matching results obtained with $f$ and $f^{s}$ in ProNet, respectively. Images with green and red borders indicate correct and wrong matching results, respectively.}
    \label{fig:retrieval_results}
\end{figure}

\noindent \textbf{Visualization of retrieval results.} 
To intuitively demonstrate the effect of class prototypes in promoting person Re-ID, we visualize the top-10 ranked retrieval results using $f$ and $f^{s}$ in ProNet, respectively. As shown in the second row of Fig.~\ref{fig:retrieval_results} (a), all six person images with the same identity as the query image are retrieved correctly at the top six of the ranking list. However, due to $f$ being the visual appearance feature extracted from the input image, some images with similar appearance but different identities are also ranked in the top six, as shown in the first row of Fig.~\ref{fig:retrieval_results} (a), and those hard samples with the same identity but with viewpoint and posture changes are not retrieved in the top-10 results.

Similarly, in the first row of Fig.~\ref{fig:retrieval_results} (b), the top one image shares similar visual characteristics with the query image, causing it to be mistakenly assumed to have the same identity as the query. However, when comparing the image feature with class prototypes, some detailed and critical identity differences can be discovered to better distinguish different people. If we take advantage of the classifier as a projection from image features to class prototypes, and use $f^{s}$ for inference, which describes the identity of input images as similarities to all prototypes, the top retrieval results are less affected by the misleading visual clues. All correct images with the same identity as the query image exhibit greater similarities. 
The retrieval results further indicate that performing Re-ID with the features projected on class prototypes shows great superiority.

In addition to enabling pedestrian images to be better retrieved, we can also discover the advantages of $f^{s}$ over $f$ from the remaining retrieval results that rank at the top. As shown in Fig.~\ref{fig:retrieval_results} (c), among the last five persons in the second row, four of them are the same person. Although due to camera changes, the illumination/color shows great differences, they are retrieved since they have similar identity characteristics with the query person. However, the last five persons in the first row just have a similar color and viewpoint to the query image.
The retrieval results confirm our claim that learned class prototypes can make a difference in identifying and distinguishing pedestrians during inference.

\section{Conclusion}
In this paper, we advocate leaving the classifier during inference to promote person Re-ID performance for the first time. We point out that the weights of the classifier can be regarded as class prototypes for effective comparison between identities, and provide new insights on identity classification loss and triplet loss to exert the positive impact of class prototypes for re-identification. With the help of class prototypes, our proposed strong baseline ProNet and multi-granularity enhanced ProNet++ achieve competitive performance without cumbersome module design, auxiliary data or annotation dependency, or heavy testing overhead. Thorough experiments demonstrate the effectiveness of class prototypes during training and inference, as well as their prospects in object re-identification. We hope that ProNet as a strong baseline can inspire more research to take advantage of the classifier, promoting Re-ID from the perspective of projection-on-prototypes in the future.

{\small
\bibliographystyle{ieee_fullname}
\bibliography{egbib}
}

\end{document}